# Single-image camera calibration with model-free distortion correction

Katia Genovese

School of Engineering, University of Basilicata, Potenza, ITALY, katia.genovese@unibas.it

**ABSTRACT**

Camera calibration is a process of paramount importance in computer vision applications that require accurate quantitative measurements. The popular method developed by Zhang relies on the use of a large number of images of a planar grid of fiducial points captured in multiple poses. Although flexible and easy to implement, Zhang's method has some limitations. The simultaneous optimization of the entire parameter set, including the coefficients of a predefined distortion model, may result in poor distortion correction at the image boundaries or in miscalculation of the intrinsic parameters, even with a reasonably small reprojection error. Indeed, applications involving image stitching (e.g. multi-camera systems) require accurate mapping of distortion up to the outermost regions of the image. Moreover, intrinsic parameters affect the accuracy of camera pose estimation, which is fundamental for applications such as vision servoing in robot navigation and automated assembly.

This paper proposes a method for estimating the complete set of calibration parameters from a single image of a planar speckle pattern covering the entire sensor. The correspondence between image points and physical points on the calibration target is obtained using Digital Image Correlation. The effective focal length and the extrinsic parameters are calculated separately after a prior evaluation of the principal point. At the end of the procedure, a dense and uniform model-free distortion map is obtained over the entire image.

Synthetic data with different noise levels were used to test the feasibility of the proposed method and to compare its metrological performance with Zhang's method. Real-world tests demonstrate the potential of the developed method to reveal aspects of the image formation that are hidden by averaging over multiple images.

1. ## INTRODUCTION

Camera calibration is the essential prerequisite for obtaining accurate metrical information from images. The image formation scheme is modelled according to the pinhole model (Fig. 1). Camera calibration consists in the estimation of the extrinsic and intrinsic parameters describing the projective transformation from 3D physical points to 2D image points. Extrinsic parameters define the pose of the camera coordinate frame $\Sigma_c$ in the world coordinate system $\Sigma_w$. Intrinsic parameters describe the affine transformation between image points in $\Sigma_c$ (in $mm$ units) to image points in the sensor frame $\Sigma_s$ (in $pixels$ units).

The literature on camera calibration is rich and extensive. Interested readers are referred to [1][2] for an overview of the subject. In the following, the analysis is limited to photogrammetric methods using a flat target with fiducial points whose position is known with high accuracy. These methods are mostly based on the technique proposed by Zhang in his seminal paper [3].

The most commonly used calibration targets consist of a regular grid of markers (circular or square) [3][4] or a checkerboard [5]. The centroids of the dots or the corners of the checkerboard serve as fiducial points in the world reference system $\Sigma_w$. Recently, the use of a speckle calibration target has been proposed [6], as a large number of fiducial points can be defined and registered with sub-pixel accuracy to their counterparts in the captured image using Digital Image Correlation (DIC) [7][8]. The inherent capabilities of the DIC method make it possible to overcome the detrimental effect that perspective deformation [4] or defocus [5] have on dot grids or checkerboards, respectively. A detailed description and discussion on the use of speckle calibration targets can be found in [6].

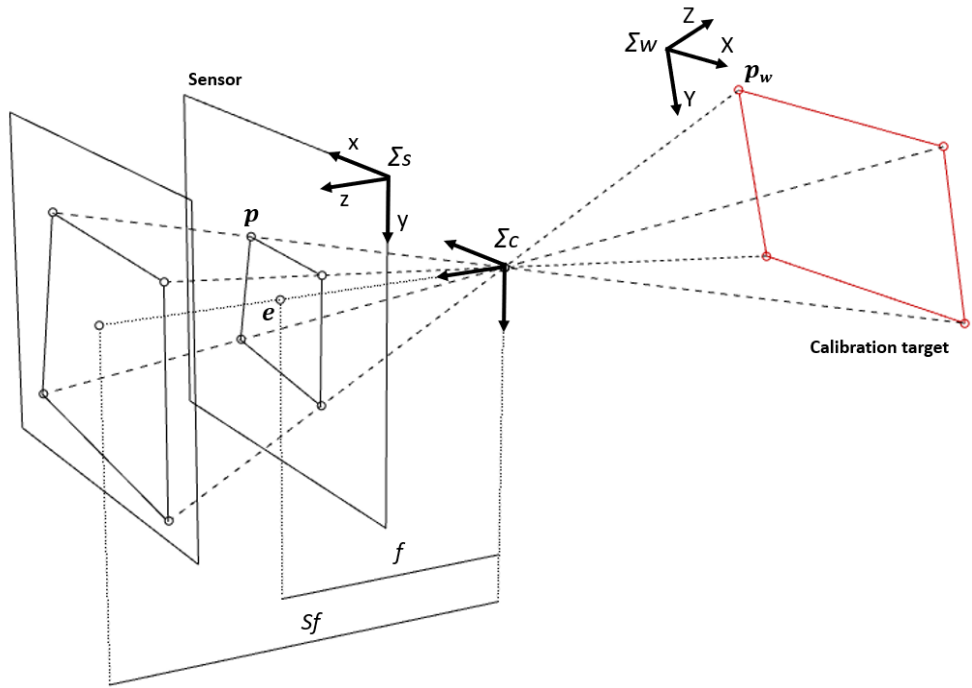

**Figure 1. Schematic of the pinhole camera model. Note how the intrinsic camera transformation is invariant to uniform scaling (with scale factor *S*) with respect to the principal point *e*.**

A further practical improvement was introduced with the use of flat digital screens to display synthetic calibration patterns. Digital screens are now ubiquitous and available in a range of sizes and spatial resolutions. Their flatness is extremely accurate and the pixel size is a readily available factory specification with precision down to $O(10^{-3})mm$. This means that once an optimal digital speckle pattern has been created using the customization possibilities offered by various methods, e.g. [9], the position in $mm$ of any point in the image domain is automatically known with great accuracy. Studies on the use of digital speckle calibration targets within the classic Zhang's method have shown superior performance compared to traditional patterns when implementing the state of the art of the DIC technique [6].

A critical issue in camera calibration is represented by image distortion, which, if not properly corrected, is a source of systematic error in video-based measurements. Most off-the-shelf camera/lens assemblies suffer from some degree of distortion which shows a dominant radial nature [10][3]. The most widely used radial distortion models are the even-order polynomial function with two or three parameters [3][11], and the division model [12], which has the advantage of leading to satisfactory results even with a single parameter. The estimation of the distortion function can be embedded in the calibration process [13][3] or it can be a stand-alone process aimed at obtaining undistorted images [14][15] and possibly the intrinsic parameters [16][17].

Zhang's method [3] belongs to the first group of calibration algorithms. A minimum of three images of the calibration target in different poses are used to compute an approximate estimate of the camera parameters via a closed-form solution based on the ideal pinhole model. These parameters are then fed as initial guess to a non-linear optimization routine which includes as design variables the coefficients of an even-order polynomial radial distortion function. The optimization routine searches for the set of parameters that minimizes the reprojection error (RPE) defined as the sum of the Euclidean distances between the real (detected) image points and their counterparts computed using the estimated perspective geometry. Although flexible and easy to implement, Zhang's method has some limitations. The number, orientation and distribution of the target poses within the camera field of view affect the calibration result. If the distortion function is estimated using points located mostly in the inner (less distorted) portion of the image, a low-order distortion function usually suffices to get low reprojection residuals. However, in this case, the distortion in the outer areas of the image may not be properly corrected and this may lead to a poor result in applications where image stitching between partially overlapping adjacent views is required (e.g., microscope images mosaicing, multi-camera DIC systems [18][19]). Conversely, if an inappropriate (e.g., over-parameterized) distortion function is forced to fit the image data, low RPE may be obtained at the expenses of incorrect (often unfeasible) values of the intrinsic parameters with direct consequences on the accuracy of the computed extrinsic parameters. In this case, there is little effect on reconstruction (e.g., for stereo DIC measurement) but a more significant effect on camera pose estimation (e.g., for visual servoing in robotics). It is widely acknowledged that the limitations above are mainly due to the interplay between intrinsic and extrinsic parameters and to the use of explicit distortion models [20][21].

To overcome the limitations associated with the use of a predefined function, some methods have been developed to estimate a parameter-free image distortion map. In their seminal paper, Hartley and Kang [22] propose a non-iterative method to compute the center of distortion ($COD$) and the radial distortion curve using a checkerboard planar pattern and a model-free approach. Although a single image is required, the method is highly sensitive to noise and the best results are obtained using multiple images [22]. Another model-free approach is presented in [15], where the distortion map, COD and effective focal length are computed using phase-shifted fringe patterns displayed on an LCD screen placed at two positions strictly parallel to the camera sensor plane. However, the method requires a high-precision translation stage, and the accuracy of the monitor orientation relative to the camera affects the calibration result. A model-free iterative distortion compensation algorithm based on phase targets is proposed in [17]. This method use a full-field information provided by multiple images of two groups of phase patterns displayed on a LCD screen and provide a full sensor distortion map. However, results are shown in terms of the RPE and no information on the accuracy of the estimated camera parameters is provided. A single pose with multiple images of structured-light patterns is used in [23] to compute a dense model-free distortion map. The method relies on the strong assumption of a null distortion at the sensor center and provide no estimation of the camera parameters. Finally, multiple images of speckle targets have been used for accurate camera calibration using model-based [6] and model-free [16] approaches. However, both approaches still need multiple poses of a speckle target each partially covering the camera sensor.

In this work, a *single* image of a speckle pattern, displayed on a monitor so as to cover the entire sensor area, is used to accurately estimate the image distortion map and the full set of calibration parameters. The potential of the method derive directly from (i) the use of a uniform and dense grid of points covering the entire image up to its boundaries, and (ii) the sub-pixel registration capabilities of Digital Image Correlation. The limitations associated with the coupling between calibration parameters are overcome by first estimating the principal point, then separately computing the effective focal length and eventually the extrinsic parameters. A final in-bundle optimization refines the calibration results by processing the obtained full-field distortion map using either a model-based or model-free approach.

The proposed Single Image Calibration (SIC) method was tested both on synthetic images with different noise levels and on real images, and its performance was compared with Zhang's method as implemented by J.Y. Bouguet in the open source Calibration Matlab Toolbox (CMT,[11]). SIC showed superior flexibility and robustness to noise compared to CMT, and captured the variation of distortion with depth, which is an aspect particularly important in close-range photogrammetry [24][25].

The paper is organized as follows. Section 2 recalls the principles of the pinhole camera model and presents the advantages of using a single image of a speckle pattern covering the entire camera sensor compared to multiple images of typical calibration grids in different poses. The SIC method is then described step by step, using a single synthetic image as an illustrative example. Section 3 reports the results of the experimental tests performed on both synthetic and real images, and compares the performance of the proposed method with Zhang's calibration technique. Due to the large amount of data generated in the study, only the most significant results are reported in the paper. Additional results are reported as online supplementary material [26], together with the full-size images of the experiments, to allow the reproducibility of the method. Finally, the advantages and limitations of SIC are discussed in Section 4, together with possible future developments and applications.

## 2. METHODS

In video-based metrology, image-formation is described by the ideal pinhole model, i.e. the image of a point and its real counterpart are related by a perspective transformation (see image formation scheme in Fig. 1). Camera calibration process consists in evaluating the parameters of the transformation from the 3D point coordinates in the world reference system $\Sigma_w$ to the 2D image point coordinates in the sensor frame $\Sigma_S$. With Zhang's method [3], camera calibration is performed using a minimum of three images of a planar grid of features (marker centroids or checkerboard corners) whose position is known with high precision. The planar calibration target is sequentially placed with different orientations within the working volume and an image of the target is captured at each pose. If the planar target is assumed to lie on $Z = 0$ of the world coordinate system, it is possible to relate the world coordinates of a target point $\boldsymbol{p_w} = [X, Y]^T$ to the coordinates of its ideal (undistorted) image projection $\boldsymbol{p} = [x, y]^T$ in the sensor plane as:

$$S \begin{bmatrix} x \\ y \\ 1 \end{bmatrix} = \begin{bmatrix} f_x & \gamma & u_0 \\ 0 & f_y & v_0 \\ 0 & 0 & 1 \end{bmatrix} \begin{bmatrix} r_{11} & r_{12} & t_x \\ r_{21} & r_{22} & t_y \\ r_{31} & r_{32} & t_z \end{bmatrix} \begin{bmatrix} X \\ Y \\ 1 \end{bmatrix} = \boldsymbol{AE} \begin{bmatrix} X \\ Y \\ 1 \end{bmatrix} \quad (1)$$

where $S$ is an arbitrary scale factor (see Fig. 1 for the geometric interpretation of the scale factor in the perspective projection model). The intrinsic matrix $\boldsymbol{A}$ defines the affine transformation between the image point coordinates from the camera reference frame $\Sigma_C$ (in $mm$ units) to the sensor image frame $\Sigma_S$ (in $pixels$ units) [3]. In particular, $f_x$ and $f_y$ are the image scale factors (also referred as focal lengths) along the horizontal and vertical directions of the sensor, $\gamma$ is the skewness of the two image axes (in this work assumed as null), and $u_0$ and $v_0$ are the coordinates of the principal point $\boldsymbol{e}$ defined as the foot of the perpendicular from the projection center to the sensor plane [3]. Intrinsic parameters are characteristics of the camera/lens assembly and independent from the pose. The extrinsic matrix $\boldsymbol{E}$ describes the rigid transformation from the world coordinate frame $\Sigma_w$ to the camera coordinate frame $\Sigma_C$. In particular, $\boldsymbol{E}$ contains the components of the first two columns $\boldsymbol{r_1}$ and $\boldsymbol{r_2}$ of the rotation matrix $\boldsymbol{R}$ and the translation vector $\boldsymbol{t} = [t_x, t_y, t_z]^T$. Using the same notation of [3], the orientation of the calibration plane is represented here by a 3D vector $\boldsymbol{r}$, which is parallel to the rotation axis and whose magnitude equals the rotation angle. For the sake of compactness, translation vector $\boldsymbol{t}$ is appended to vector $\boldsymbol{r}$, to give the extrinsic parameters vector $\boldsymbol{V_E} = [\theta_x, \theta_y, \theta_z, t_x, t_y, t_z]^T$. The intrinsic parameters will be stored in the vector $\boldsymbol{V_I} = [f_x, f_y, u_0, v_0]^T$.

The homography $\boldsymbol{H} = \boldsymbol{AE}$, which describes the perspective transformation between a physical point $\boldsymbol{p_w}$ on the model plane and its image $\boldsymbol{p}$ in the sensor plane, can be estimated, up to a scale factor, from a single image of the model plane. The procedure for its estimation, which requires at least four fiducial points, is described in detail in Appendix A of reference [3]. However, to individually extract the intrinsic and extrinsic parameters from the nine components of $\boldsymbol{H}$, at least $n = 3$ images (or $n = 2$ if $\gamma = 0$) of the calibration plane in different poses, and thus $n$ different homographies, are required. It has been shown that using a larger number of calibration pattern poses, typically $n \geq 10$, improves both the accuracy and the precision of the method [3].

The presence of distortion in real images is addressed by defining an appropriate function that relates the coordinates of the detected image point $\boldsymbol{p_d} = [x_d, y_d]^T$ to the ideal point coordinates $\boldsymbol{p}$ calculated with (1). Although various sophisticated distortion models have been proposed in the literature [24], it is generally accepted that the radial component of the distortion is dominant over the other components. Adding further components to the distortion function is often irrelevant in terms of RPE, while it may negatively affect the feasibility of the results. In this work, only the radial component of the distortion is considered and, similar to CMT, it is assumed to be described by an even order polynomial function of 6$^{th}$-order as follows:

$$x_d = f_x(x_n + x_n[k_1 r_n^2 + k_2 r_n^4 + k_3 r_n^6]) + u_o$$
$$y_d = f_y(y_n + y_n[k_1 r_n^2 + k_2 r_n^4 + k_3 r_n^6]) + v_o \quad (2)$$

where:

$$r_n^2 = x_n^2 + y_n^2 \quad (3)$$

and the normalized ideal coordinates are calculated as:

$$x_n = (x - u_o)/f_x$$
$$y_n = (y - v_o)/f_y. \quad (4)$$

The coefficients $k_1, k_2$ and $k_3$ represent the parameters of the radial distortion function. Similarly to [3], the center of distortion is assumed to coincide with the principal point $\boldsymbol{e}$. When a prescribed analytical model is used to define the deviation of the distorted points $\boldsymbol{p_d}$ from the position of the ideal projected points $\boldsymbol{p}$, the distortion parameters are added as unknowns to the set of intrinsic parameters. Successful distortion removal results in the retrieval of undistorted points $\boldsymbol{p}$, which are related to the target points $\boldsymbol{p_w}$ by the ideal homography $\boldsymbol{H}$ (Eq.1). It is important to note that there exists an ambiguity between the amount of distortion and the scale factor (focal length parameter) [22]. This ambiguity is typically solved by assuming that the radial distortion is null at the distortion center where distorted and undistorted points coincide.

In Zhang's method, the presence of distortion is handled with a two-step approach. In a first step, the images are assumed to be free of noise and distortion, and an analytical solution is used to compute an approximate estimate of the intrinsic and extrinsic parameters. The obtained values are then used as an initial guess in a non-linear optimization problem that refines the entire set of parameters, including the coefficients of the distortion function. The objective function is represented by the reprojection error RPE, i.e. the Euclidean distance between the distorted image points (as calculated by Eqs.(1)-(4)) and their detected counterparts. For highly distorted images, the analytical calculation of the principal point and focal length [3] may fail. For this reason, alternative approaches are often used to calculate their approximate initial value. For example, in the CMT [11], which is considered in this paper for comparison with the proposed method, the principal point is assumed to be at the center of the sensor and the focal length is estimated from vanishing points.

Although Zhang's method is undoubtedly powerful and easy to implement, it suffers from some limitations:
- A sufficiently large number of images is required to reduce the uncertainty of the results (typically $n \geq 10$).
- The position and orientation of the calibration target within the working volume affect the calibration results.
- Perspective deformation can affect the accuracy of the centroid location for the (circular or square) markers when using a calibration dot pattern; defocusing can affect the accuracy of corner location in the case of a checkerboard pattern.
- A prescribed distortion function is forced to adapt to the real distortion of the lens/camera assembly. Therefore, especially when implementing a large number of distortion parameters, the results of the final optimization may not accurately reflect the ground-truth even at low RPE values.
- Due to the coupling between the design parameters and the non-linearity of the optimization problem, inaccurate initial values of the parameters and inappropriate bounds can potentially lead to convergence to a local minimum. The intrinsic parameters are particularly susceptible to estimation errors.

This work aims to overcome the above limitations as follows:
- A speckle distribution is used as a calibration pattern, which allows the point pair $\boldsymbol{p_d}$ and $\boldsymbol{p_w}$ to be matched with subpixel accuracy using DIC. Furthermore, a speckle pattern is robust to both perspective distortion and defocus [27]. The advantages of using a speckle calibration pattern over a traditional dot/checkerboard pattern have been presented and discussed in [6] and will not be repeated here.
- A single image is sufficient to remove distortion and compute an accurate initial estimate (with less than 1% error with respect to the ground truth) for the full set of calibration parameters. The only requirement is to achieve full coverage of the sensor with the speckle image (see Fig.2d). This is the main novelty of the method presented in this work.
- An accurate estimation of the center of distortion and of the focal length is performed separately from the other design variables, thus avoiding coupling errors.
- Image distortion can be densely and uniformly mapped over the entire sensor area up to its boundaries. This map can either be used as a model-free pointwise warping function or interpolated with an analytical model deemed appropriate for the current lens/camera system.

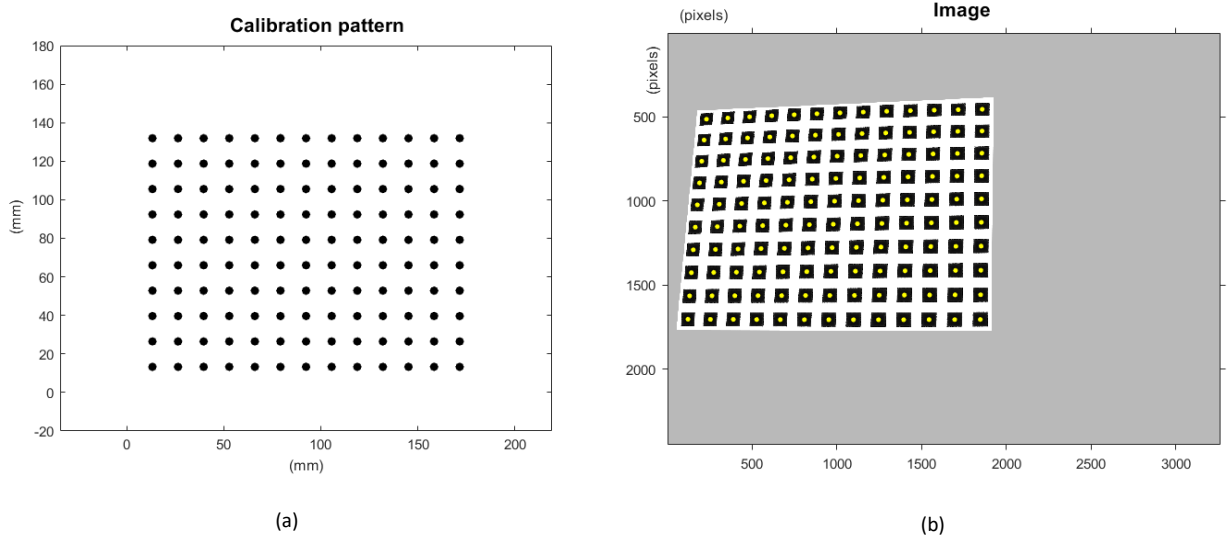

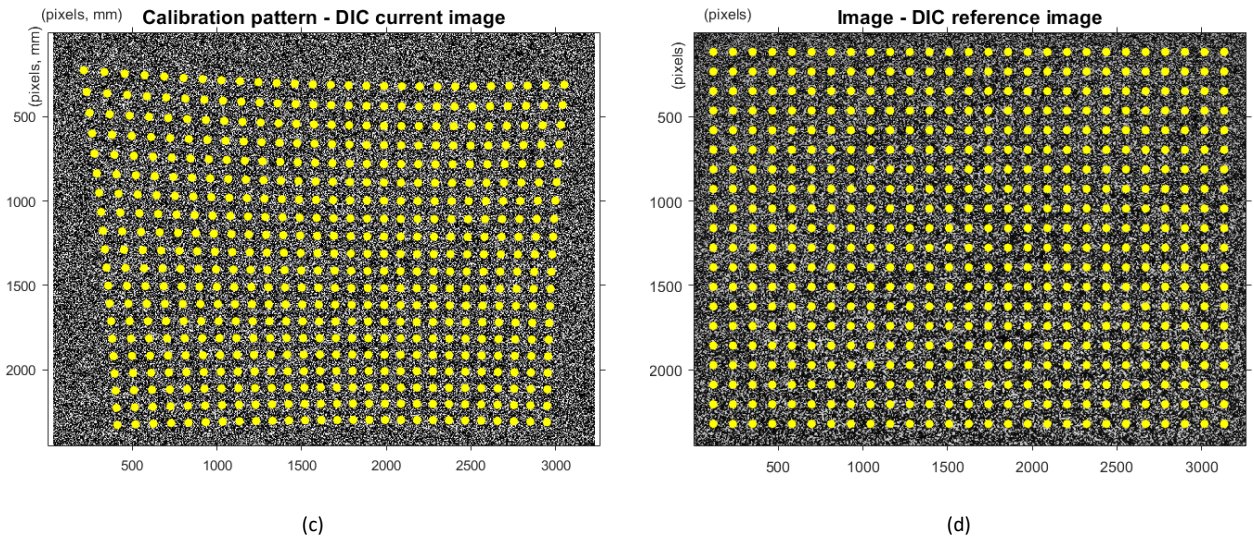

**Figure 2.** Different use of the calibration pattern in Zhang's method and in the proposed method. In Zhang's method, a large number of images of a flat calibration target with a grid of fiducial points (a) are captured in different poses, e.g. (b). In the SIC method, a synthetic image of a random pattern (c) is displayed on a monitor with a known pixels/mm ratio and captured by the camera so as to cover the entire sensor area (d). A regular grid of points is then defined on the captured image (reference image) and the corresponding grid on the synthetic image (current image) is matched using DIC to calculate the position of the corresponding target points in mm.

The above properties are a direct consequence of using a single image of the speckle calibration target that covers the entire area of the sensor. Figure 2 illustrates the fundamental difference between the conventional Zhang's camera calibration method (ZM) and the approach proposed in this work. In ZM, a grid of fiducial points $p_w$ is defined in a reference system attached to the model plane (Fig.2a), and the position of the corresponding image points $p_d$ is calculated in each pose (Fig.2b)[1]. In the SIC method, a dense and regular grid $p_d$ is defined over the entire area of the captured image (Fig.2d), and the corresponding target points $p_w$ on the monitor are matched by DIC on the synthetic calibration image for which the conversion factor from $pixels$ to $mm$ is known with high accuracy (Fig.2c). Thus, in the latter case, the point grid $p_d$ covers the entire image with a high and uniform density, (in Fig.2d, only a coarse point grid is shown for clarity), while the target points are unevenly distributed over an irregular area of the synthetic speckle image (Fig.2c). On the contrary, for ZM, the distorted (by perspective) image of the regular grid of points in the model plane (Fig.2a) covers unevenly only a portion of the image (Fig.2b). Furthermore, the number $n_p$ of points in the calibration grid differs significantly between the two cases. For a standard calibration target, $n_p$ is of the order of hundreds (in this work $n_p = 130$). For the speckle target, $n_p$ depends on the image size and the parameters of the DIC analysis (subset and spacing), but it is about $10^3$ times larger than for the standard calibration grid (in the simulated images of this work $n_p = 126505$ points).

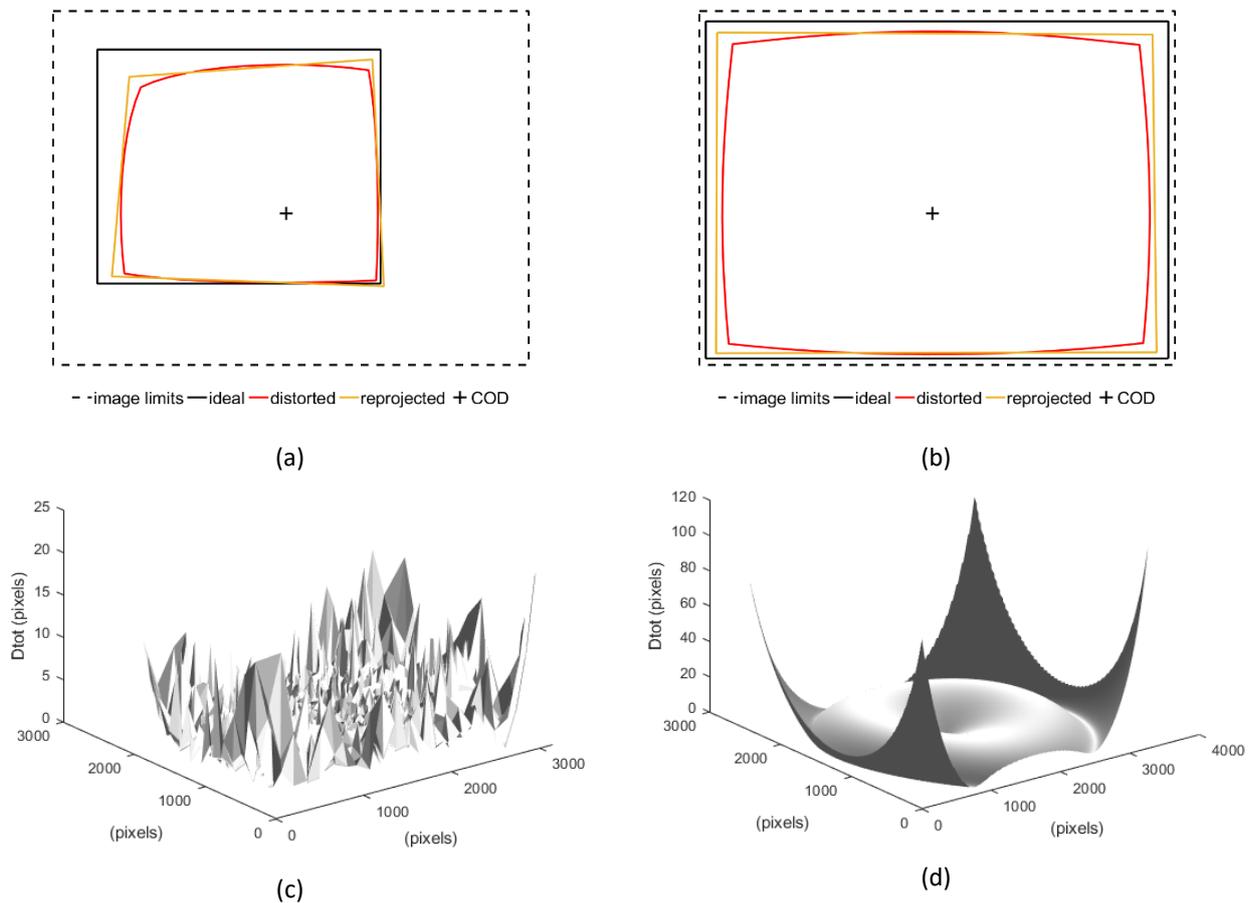

Figure 3. Synthetic images illustrating the effect that distortion and a decentered COD have on the homography $H_d$ calculated from the distorted points of a planar calibration target placed parallel to the sensor. In (a) the image of the target covers only part of the image, while in (b) it occupies the entire sensor area. Note that, in case (a), the reprojected points $p_p$ obtained from $H_d$ clearly lie on a plane tilted with respect to the plane of the sensor, thus revealing an erroneously estimated homography associated with the projection. On the contrary, in case (b), the reprojected points $p_p$ appear as the result of an affine transformation applied to the ideal points $p$. Distribution of total disparity $D_{tot}$ for simulated data of a set of 20 poses of a planar pattern with a grid of 130 points (c). Distribution of total disparity $D_{tot}$ for a single simulated image of a pattern speckle covering the entire area of the sensor (d). Note the different scale for the two distributions $D_{tot}$.

---

[1] For brevity, the suffix $i = 1, \ldots n_p$ - where $n_p$ is the number of points of the calibration grid- is omitted for the generic physical point $p_w$ and corresponding image point.

The different definitions of the $p_w - p_d$ point pair have a dramatic effect on the result of the homography calculation from the distorted image points. It is important to note that the reprojected points $p_p = [x_p, y_p]^T$ calculated as $p_p = Hp_w$ (Eq.1) coincide with $p$ only for ideal lenses. When the homography is computed from the real (distorted) image points $p_d$, it describes the perspective transformation of $p_w$ that generates the maximum likelihood estimate of $p_d$. Importantly, the homography $H_d$ (the suffix $d$ is used to distinguish it from the ground truth $H$) is computed by considering the distorted points with the same weight regardless of their distance from the center of the distortion. Figure 3 illustrates the different results of the calculation of $H_d$ for an image acquired according to ZM (Fig.2b) and for a point grid covering the entire image area as required by SIC (Fig.2d). In particular, Figure 3 refers to synthetic images generated with $R = I$ (i.e., with the model plane parallel to the sensor) to highlight the effect of the sole distortion on the reprojected point grid. For illustrative purposes, the synthetic images have a strong barrel distortion and a highly decentered COD. For clarity, only the boundaries of the rectangular point grid domain is shown in the figure.

By comparing the two simulated images, it can be seen that the reprojected points in Fig.3a differ greatly from the ground truth as a direct consequence of three factors: (i) the partial coverage of the sensor, (ii) the inhomogeneity of the point spacing introduced by the distortion, and (iii) the decentering of the COD. In particular, the reprojected point domain is shifted towards the COD and, more importantly, it appears tilted with respect to the ideal reprojected plane (parallel to the sensor), thus revealing an erroneously estimated homography associated with it. On the contrary, in Fig.3b, the reprojected points appear as a mere scaled copy of the ground truth. Indeed, there are also an anisotropy in the scale factor (due to the different number of points in the horizontal and vertical directions of a rectangular sensor) and a slight drift of the pattern towards the COD. For a given level of distortion, the similarity between the reprojected pattern and the ground truth in Fig.3b is as close as the number of points considered is large. The same considerations apply also to the case of $R \neq I$. The existence of an (approximate) affine transformation between the reprojected points and the ground truth (Figs.3b,d) is at the basis of the calibration method developed in this work.

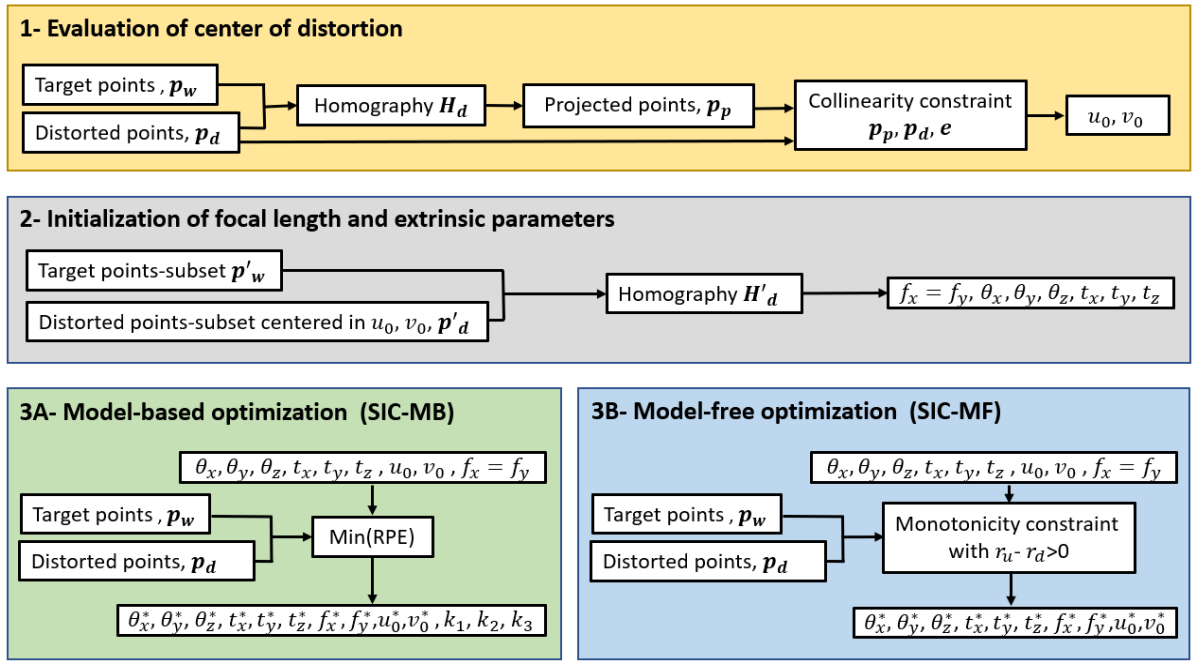

Figure 4. Workflow of the Single Image Calibration (SIC) method developed in this work. Optimal calibration parameters are marked with (*).

Figures 3c and 3d report the total disparity $D_{tot} = \|p_p - p_d\|$ between the distorted and reprojected synthetic noise-free data for a set of 20 poses of a standard calibration target (see Fig.1 in [26]) and for a single pose of a speckle target, respectively. In the first case, any $n^{th}$ estimated homography $H_d$ generates a set of reprojected points $p_p$ that represent the best *local* approximation of the image grid points $p_d$ in the $n^{th}$ pose (i.e. $p_p \approx p_d$, hence the low $D_{tot}$ value), but it gives an inaccurate contribution at the global level ($p_p \neq p$). On the contrary, the plot in Fig.3d clearly shows that a first close (scaled) approximation of the *global* distortion distribution (i.e. $p_p \approx Sp$, hence the high $D_{tot}$ value at the edges of the image) can already be obtained from the $H_d$ estimated from a single speckle image covering the entire sensor.

Figure 4 shows the workflow of the Single Image Calibration (SIC) method. The first two steps of the procedure are aimed at obtaining a scaled close approximation of the undistorted image points, from which a very accurate initial guess (error < 1%) of the full set of intrinsic and extrinsic parameters is computed. The calibration can then be refined either by defining an analytical distortion function (Step #3A, SIC-*Model-Based* approach, SIC-MB) or, alternatively, by using the computed dense and uniform distribution of the radial distortion data to perform a pointwise unwarping of the image (Step #3B, SIC-*Model-Free* approach, SIC-MF).

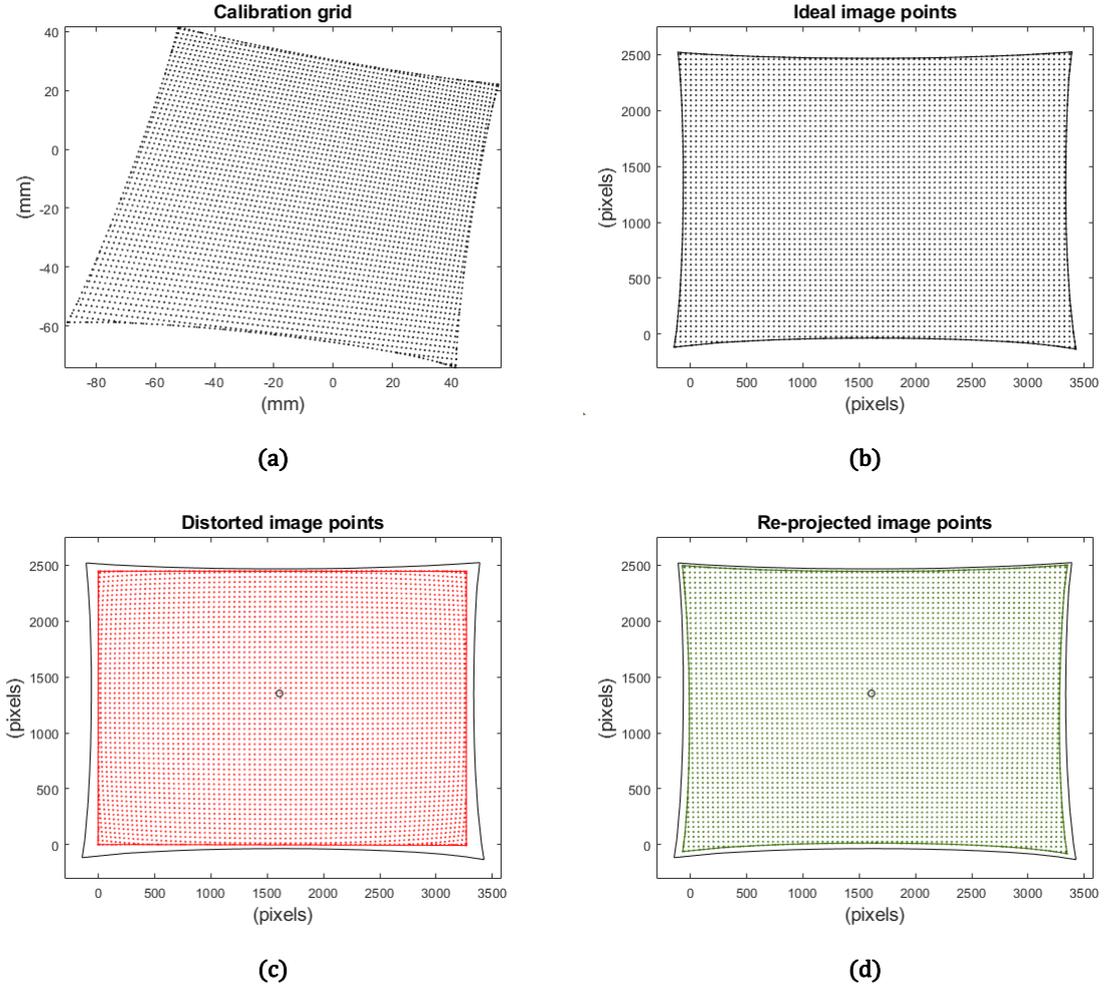

**Figure 5.** Synthetic data simulating the DIC point grid defined over the image of a speckle pattern covering the entire sensor area (delimited by a continuous red line in panel (c)). Calibration grid $p_w$ (a), ideal image points $p$ (b), distorted (detected) points $p_d$ (c) and reprojected points $p_p$ (d). The boundaries of the ideal points domain and the COD are also shown in (c) and (d) for better comparison. For clarity, only 4755 points of the 126505 simulated image points are plotted.

To better describe step by step the proposed procedure, the pose #1 of the simulated data later reported in Section 3 has been considered as an illustrative example. The pose is characterized by all non-null extrinsic parameters ($V_E = [8°, 16°, -26°, 5, 8, 300]^T$) and thus it is representative of a general case (see Fig.2 in [26] and illustrative coarse grids in Figs.5a,b). The intrinsic parameters and the coefficients of a barrel distortion function were set as $V_I = [9285.7, 9278.6, 1609, 1353]^T$ and $K = [-1.3, 8.8, -163]^T$, respectively. The center of distortion (coinciding with the principal point $e$) has been deliberately set significantly away from the center of the sensor $e_s = [u_s, v_s]^T = [1632, 1224]^T$. The image data were generated by projecting a dense regular grid of $n_p = 126505$ target points $p_w$ over the entire image, simulating the point data $p_d$ of the speckle image required for the SIC method (Fig.2d). Figure 5d shows the boundaries of the ideal point domain - with a continuous black line - and the reprojected points $p_p$ obtained from the homography $H_d$ calculated from the distorted points $p_d$ (shown in Fig.5c).

The steps required to calibrate the camera from a single image (Fig.5c) are detailed below.

*Step #1: Evaluation of the center of distortion.*

Figure 5d shows that when a very dense grid of points $p_d$ covering the entire image is used to retrieve the homography $H_d$, the reprojected points $p_p$ appear to differ from the ground truth $p$ by only a scale factor $S$. Indeed, the reprojected points $p_p$ do not coincide with any set of undistorted points $p_u = [x_u, y_u]^T = Sp$ as the calculation of the homography $H_d$ is influenced by the different number of points $p_d$ in the horizontal and vertical directions (in the most common case of a rectangular sensor) and, along the same direction, by the different degree of distortion of the pattern due to the $COD$ eccentricity. Therefore, with respect to the undistorted points $p_u$, the $p_p$ distribution will have an aspect ratio $AR = S_y/S_x \neq 1$ (where $S_x$ and $S_y$ are the scale factors along the horizontal and vertical directions) and will be shifted towards (away from) the $COD$ for barrel (pincushion) distortion. With sufficient approximation, it can be hence assumed that there is an affine transformation between the reprojected points $p_p$ and the undistorted points $p_u$.

To evaluate this transformation, the two sets of points $p_d$ and $p_p$ are first centered to the origin of the sensor frame by shifting them by $(u_0, v_0)$ ($COD$ coordinates) and $(u_c, v_c)$ (coordinates of the center of the reprojected point pattern), respectively:

$$p_{d0} = \begin{bmatrix} x_{d0} \\ y_{d0} \end{bmatrix} = \begin{bmatrix} x_d - u_0 \\ y_d - v_0 \end{bmatrix}$$
$$p_{p0} = \begin{bmatrix} x_{p0} \\ y_{p0} \end{bmatrix} = \begin{bmatrix} x_p - u_c \\ y_p - v_c \end{bmatrix}. \tag{5}$$

Then, anisotropic scaling is applied to the centered projected points $p_{p0}$ to obtain the centered undistorted points $p_{u0}$ as follows:

$$p_{u0} = \begin{bmatrix} x_{u0} \\ y_{u0} \end{bmatrix} = \begin{bmatrix} S_x & 0 \\ 0 & S_y \end{bmatrix} \begin{bmatrix} x_p - u_c \\ y_p - v_c \end{bmatrix} = \begin{bmatrix} x_u - u_0 \\ y_u - v_0 \end{bmatrix}. \tag{6}$$

Since only radial distortion is considered, each pair of points $p_{d0}$ - $p_{u0}$ must lie on a line $\ell$ passing through the origin of the sensor frame. The set of the unknown parameters in Eq.(6) can therefore be estimated by imposing the collinearity constraint, i.e. by minimizing the sum $CC$ of the distances of the $n_p$ lines $\ell$ from $(0,0)$, defined as follows:

$$CC(S_x, S_y, u_0, v_0, u_c, v_c) = \frac{1}{n_P} \sum_{i=1}^{n_P} \frac{|x_{u0}(i) y_{d0}(i) - y_{u0}(i) x_{d0}(i)|}{\sqrt{(y_{d0}(i) - y_{u0}(i))^2 + (x_{d0}(i) - x_{u0}(i))^2}}. \tag{7}$$

A robust initial guess for the affine transformation parameters is $S_x = S_y = 1$, $u_0 = u_c = x_{min}$ and $v_0 = v_c = y_{min}$ where $(x_{min}, y_{min})$ is the minimum in the central region of the total disparity map $D_{tot}(x_d, y_d)$ (see Fig.3d). In this first step, it is important to set large bounds for scale factors $S_x$ and $S_y$. For all synthetic and real data tested in this study, the non-linear optimization process converges successfully even assuming $(x_{min}, y_{min}) = (u_s, v_s)$. On the contrary, it is important to keep the variables $u_0, v_0$ separate from $u_c, v_c$.

The computed optimal position of the $COD = [u_0, v_0]^T$ differs from the ground truth by less than $1\ pixel$ (Table 1, noiseless data), even in the presence of significant noise (see Table 1 in [26]). The remaining parameters of the estimated set of design variables are not needed for the rest of the procedure. It is interesting to note that, although the computed aspect ratio is very close to unity ($AR = 1.0048$), the two total disparity maps $D_{tot\_p} = \|p_p - p_d\|$ (Fig.6a) and $D_{tot\_u} = \|p_u - p_d\|$ (Fig.6b) differ significantly. Comparing the plots in Fig. 6d and Fig. 6e, it is possible to note that the collinearity constraint between $p_d$, $p_u$ and $COD$ is now satisfied (the curve $r_u - r_d$ passes through (0,0), where $r_u$ and $r_d$ are the distances of the undistorted and distorted points from the $COD$, respectively). However, the radial distortion function $r_u - r_d$ is still not circularly symmetric (Fig.6b). In other words, at the end of this step it still holds that $p_u \neq Sp$.

Table 1. Computed intrinsic and extrinsic parameters for a single noise-free simulated image at different steps of the SIC method.

| Method | $u_0$ (pixels) | $v_0$ (pixels) | $f_x$ (pixels) | $f_y$ (pixels) | $\theta_x$ (°) | $\theta_y$ (°) | $\theta_z$ (°) | $t_x$ (mm) | $t_y$ (mm) | $t_z$ (mm) |
|---|---|---|---|---|---|---|---|---|---|---|
| Ground truth | 1609 | 1353 | 9285.7 | 9278.6 | 8 | 16 | -26 | 5 | 8 | 300 |
| SIC-Step #1 | 1609.08 | 1352.7 | 8186.91 | 8186.91 | 7.99 | 15.25 | -27.12 | 5.01 | 8.03 | 272.90 |
| SIC-Step #2 | 1609.08 | 1352.7 | 9093.62 | 9093.62 | 7.99 | 16.01 | -26.03 | 4.99 | 7.99 | 296.95 |
| SIC-Step #3A | 1608.93 | 1352.94 | 9285.28 | 9278.04 | 7.99 | 16.00 | -25.99 | 5.00 | 8.00 | 299.98 |
| SIC-Step #3B | 1609.07 | 1352.73 | 9284.34 | 9277.50 | 7.99 | 15.99 | -26.00 | 4.99 | 8.00 | 299.98 |

The optimization procedure described above for evaluating the $COD$ strictly requires a highly dense grid of points distributed over the entire image (Fig.2, [26]). When applied to the full set of reprojected and distorted points of the 20 images used for ZM (Fig.3c), it fails, even when the ground truth is fed as an initial guess.

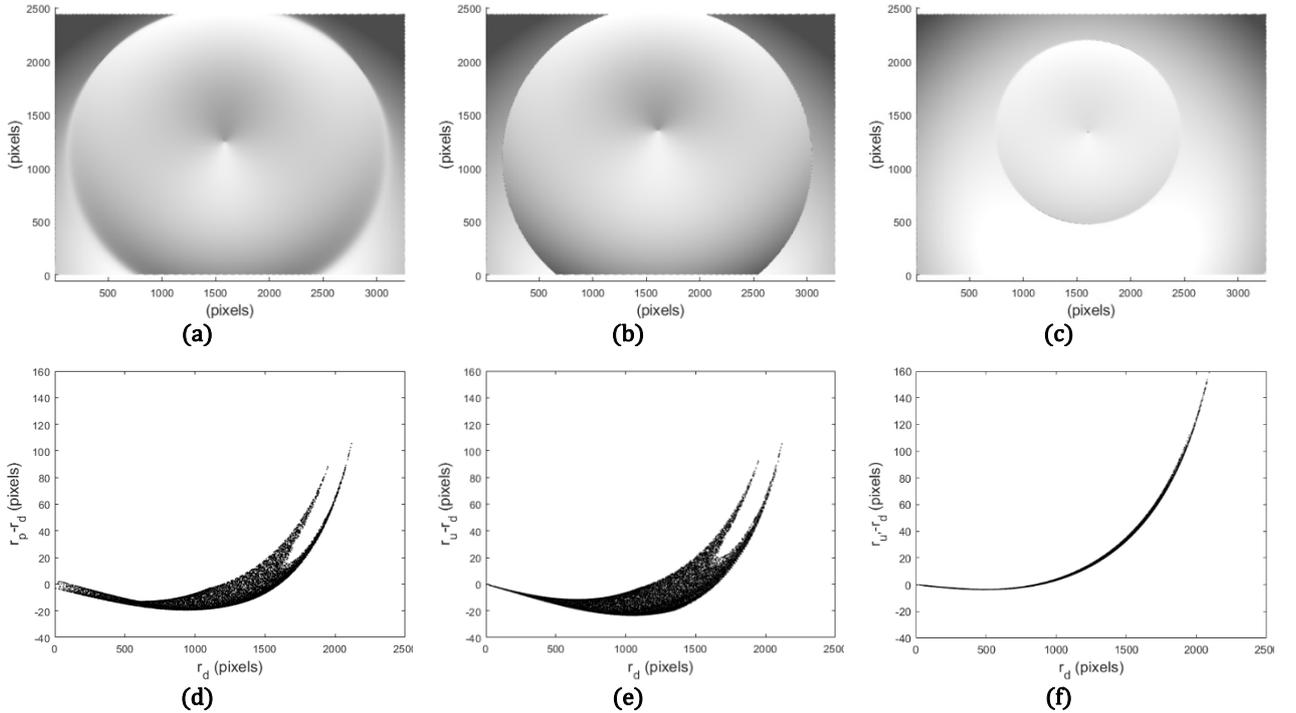

Figure 6. Top view of the total disparity maps $D_{tot\_p}$ (a), $D_{tot\_u}$ (b) and $D_{tot\_p'}$ (c) for the noise-free simulated data for pose #1 (z-axis as in Fig. 3d). Corresponding radial distortion curves at the start (d) and at the end (e) of the Step #1, and at the end of Step #2. At the end of Step #1 all lines connecting $p_u$ with $p_d$ pass through the $COD$ (point (0,0) in the plot) but the distortion function is not circularly symmetric (e). At the end of Step #2 a radial circularly symmetric distortion function is obtained which differs from the ideal curve only by a scale factor (f).

*Step #2: Initialization of focal length and extrinsic parameters.*

Once an accurate initial estimate of the coordinates of the center of distortion has been computed, it is possible to refine the homography computation by considering the subset $\boldsymbol{p'}_d$ of the distorted points (and the corresponding subset of target points $\boldsymbol{p'}_w$) that lie within the maximum circle centered in $(u_0, v_0)$ that is fully contained in the image. In this way, the number of points with the same level of distortion is fairly equal in all directions. In fact, as a proof, if the optimization problem of Eq.(7) is solved for this points subset, it results in an aspect ratio $AR$ different from unity by $10^{-5}$, $u_c = u_0$ and $v_c = v_0$.

If $h_{jk}$ (with $j = 1,2,3$ and $k = 1,2,3$) are the components of the homography $\boldsymbol{H'}_d$ relating the points $\boldsymbol{p'}_d$ to $\boldsymbol{p'}_w$, a first estimate of the focal lengths $f_x$ and $f_y$ can be calculated as follows:

$$f_x = f_y = \sqrt{\left|\frac{(h_{11}-u_0 \cdot h_{31})(h_{12}-u_0 \cdot h_{32})+(h_{21}-v_0 \cdot h_{31})(h_{22}-v_0 \cdot h_{32})}{h_{31} \cdot h_{32}}\right|}. \tag{8}$$

This formula only holds if $h_{31} \cdot h_{32} \neq 0$. However, since $h_{31}$ and $h_{32}$ are proportional to $r_{31}$ and $r_{32}$, it is easy to calculate and discard the ill-posed configurations. For example, it is important to avoid placing the calibration target parallel to the sensor even if, interestingly, '*It is alarming news that perpendicular views to planar objects are most common both in final applications as well as in camera calibration*' [28].

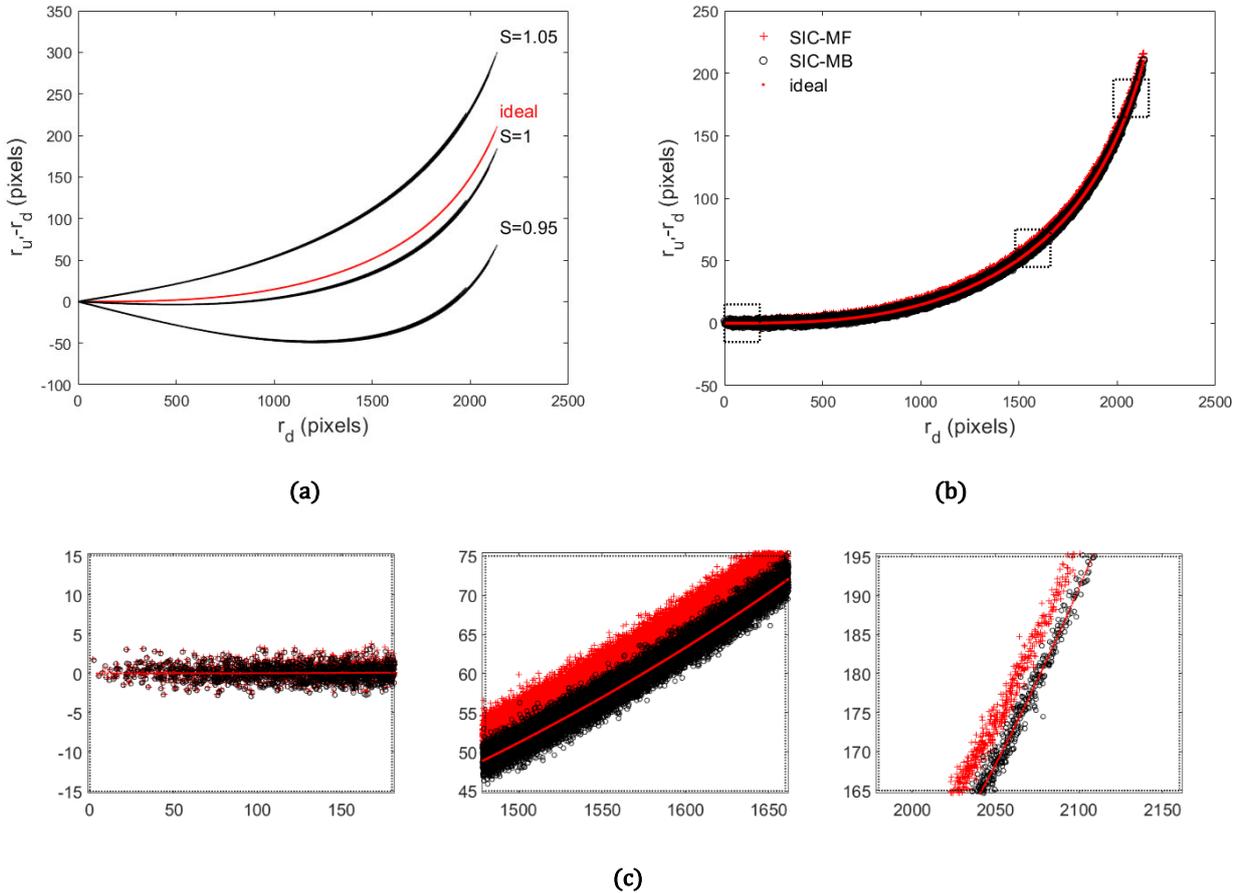

**Figure 7.** Radial distortion curves $r_{u'} - r_d$ as obtained at the end of Step#2 from points $\boldsymbol{p'}_u$ multiplied by different scaling factors $S$ compared to the ideal radial distortion curve obtained from the noise-free ideal coordinates $p$ (a). Radial distortion curves obtained with the two SIC approaches for simulated data with the highest noise level tested ($\sigma = 1\ pixel$) (b). Enlarged views of different regions of the curves in (b) are shows in the insets of panel (c).

At the end of this step, a first accurate estimate of the intrinsic matrix $\boldsymbol{A}$ is calculated, from which the extrinsic parameters can be obtained [3]. Figure 6c shows the total disparity map $D_{tot\_u'} = \|\boldsymbol{p'}_u - \boldsymbol{p}_d\|$ obtained with the updated undistorted points $\boldsymbol{p'}_u = \boldsymbol{H'}_d\ \boldsymbol{p}_w$. A circularly symmetric distortion curve $r_{u'} - r_d$ is computed

(where $r_{u\prime}$ is the distance of $p'_u$ from the $COD$, Fig.6f) which is a fairly accurate *scaled* estimate of the ideal points $p$, i.e. $p'_u \approx Sp$ (see Fig.1 for the representation of two differently scaled projections of $p_w$). This can also be inferred from Table 1, which shows a very small error between the ground truth and the parameter values estimated at the end of Step #2. Only the estimated focal length is smaller than the true value because the reprojected points best fit a point set with a barrel distortion. However, since the homography $H'_d$ has been calculated without considering the most marginal points of the image, the scaling factor between $p'_u$ and $p$ is very close to unity. This means that, for undemanding applications, it is possible to calibrate the camera from a single image of a speckle pattern and obtain a fairly accurate set of calibration parameters, as well as an undistorted image, without the need to define a distortion model.

Figure 7a shows the radial distortion curves obtained by multiplying $p'_u$ by different scaling factors $S$ compared to the ideal radial distortion curve. The scale factor is related to the value of the focal length and affects the parameters of the distortion function [22] (Fig.1). The ideal distortion curve is monotonous with zero value and zero slope at the center of distortion [22]. Starting from the obtained radial distortion curve, a further refinement step aims at finding the homography that leads to the ideal radial distortion curve (red curve in Fig.7) using two alternative optimization approaches.

*Step #3A: Model-based optimization (SIC-MB).*

In this step, common to most calibration methods, the parameters obtained in the previous step are refined together with the parameters of a prescribed distortion function (Eq.2) by minimizing the reprojection error. The non-linear optimization problem is solved here using the Sequential Quadratic Programming algorithm implemented in Matlab. The initial guess for the distortion parameters $k_1, k_2$ and $k_3$ is set to zero. The very accurate initial guess obtained in the previous step, together with strict bounds, allows to increase the computational efficiency of the optimization process and to avoid the drift towards unfeasible local minima. This is especially the case for the principal point, which is particularly sensitive to noise and to any possible misfit of the real data to the prescribed function, given the rather flat distribution of the distortion function around the $COD$ (see Fig.7a). The calculated RPE for the noise-free synthetic image tested in this Section is $1.2 \cdot 10^{-3} \pm 1.6 \cdot 10^{-3}\ pixels$. The optimized values of the calibration parameters are given in Table 1. The evaluated values of the distortion parameters are $k_1 = -1.3, k_1 = 8.81, k_1 = -163.18$. The resulting final radial distortion curve (not shown) overlaps the ideal curve in Figure 7a with no noticeable difference.

*Step #3B: Model-free optimization (SIC-MF).*

The highly dense and uniform grid of image points considered in the SIC procedure results in a full-field mapping of the distortion over the entire image, thus allowing a model-free approach to distortion removal. The radial distortion curve obtained at the end of Step #2 (Fig.6f) can be further refined without forcing it into a prescribed analytical function by implementing an approach similar to that described by Hartley and Kang in [22]. In particular, starting from the initial estimate obtained at the end of Step #2, the calibration parameters are optimized by imposing the monotonicity of the curve $r_{u\prime}(r_d)$, i.e. by imposing that the radial distance $r_{u\prime}$ of the undistorted points $p'_u$ from the $COD$ is a monotonic function of the radial distance $r_d$ of their distorted counterparts $p_d$ (i.e. the order of the points remains unchanged). This is obtained by first sorting the distorted radii $r_d$ in ascending order, then rearranging the undistorted radii $r_{u\prime}$ accordingly, and finally performing a non-linear search to minimize the following function:

$$M(u_0, v_0, FR, \theta_x, \theta_y, \theta_z, t_x, t_y, t_z) = \sum_{i=1}^{n_p-1}(r_{u\prime}(i+1) - r_{u\prime}(i))^2. \tag{9}$$

Due to the ambiguity between the scale of the distortion and the scale of the homography, the focal length $f_x$ is kept constant and equal to the value calculated in Step #2, while the ratio $FR = f_y/f_x$ is included as a design variable of the optimization. As a result of this step, the dispersion of the radial distortion curve at $S = 1$ in Fig.7a is significantly reduced while maintaining its original scale.

Finally, the optimal scale factor $S^*$ ( with $S^* \boldsymbol{p'}_u = \boldsymbol{p}$ and $S^* f_x$ the focal length corresponding to the ground truth) is found by minimizing the following function:

$$O(S) = \sum_{i=1}^{n_p}\left(Sr_{u'}(i) - r_d(i)\right)^2 \tag{10}$$

with the constraint $Sr_u(i) - r_d(i) > -\varepsilon$ with $\varepsilon \geq 0$. In other words, among all the possible monotonous curves in Fig. 7a, the curve tangent to the horizontal axis is sought. Theoretically, $\varepsilon = 0$. In practical cases, $\varepsilon$ should be chosen according to the (unknown) noise level. For synthetic data, since the value of the noise standard deviation $\sigma$ is known, $\varepsilon$ was chosen equal to be $3.5\sigma$. For experimental image data, an arbitrary choice of $\varepsilon$ would affect the accuracy of the focal length evaluation. Therefore, an alternative approach has been implemented that is robust to noise. It consists in minimizing the function $O(S)$ (Eq.10) with the constraint $median(Sr_{u'}(i) - r_d(i)) = 0$ for the first $n'_p$ points closest to the $COD$. The optimal value of $n'_p$ is defined by the user depending on the spacing of the DIC point grid and the size of the image. For the data reported in this paper, $n'_p$ was set equal to 200. Even in the case of the model-free approach applied to noise-free data, the resulting final radial distortion curve overlaps the ideal curve in Figure 7a with no noticeable difference (plot not shown).

The distortion curve obtained by either the model-based or the model-free method can be used to correct any other image captured with the same camera. Indeed, we show in the Section 3.2 that the assumption of constancy of the distortion function within the working volume may not be a safe assumption in the case of real images.

### 3. EXPERIMENTS

*3.1 Synthetic image data*

To further validate the proposed calibration method and compare its performance with Zhang's method, synthetic image data corresponding to 20 different poses of a standard calibration target were generated by projecting a grid of $13 \times 10$ points with a pitch of $5.28\ mm$ onto a $3264 \times 2448$ pixels sensor. The full set of image data tested, together with the corresponding values for $V_E$, $V_I$ and $K$ is shown in Fig.1 of the supplementary material [26]. The simulated image data have been processed with CMT [11] for ten levels of Gaussian noise with standard deviation $\sigma$ in the range $[0,1]\ pixels$.

The intrinsic parameters obtained with the three approaches CMT, SIC-MB and SIC-MF are compared in Figs.8a-d as a function of the noise level. For noiseless image data, since CMT implements the exact distortion function in Eq.2, the CMT results coincide with the ground truth. For all non-zero noise levels, the SIC-MB method outperforms the other two approaches. In particular, the SIC-MB method appears to be comparatively insensitive to noise, especially with respect to the estimation of the distortion coefficients (Fig.8e). For $\sigma > 0$, the SIC-MB method also outperforms the CMT method in terms of the total disparity between the ground truth and the undistorted points (Fig.8f). The lack of a clear trend for the SIC-MF depends on the choice of $\varepsilon$, which makes the calibration result very sensitive to outliers. The effect of noise on the estimation of the radial distortion curve for the SIC method is shown in Fig.7b for the simulated data with the higher level of noise tested ($\sigma = 1\ pixel$). The ideal curve appears as the best fit of the SIC-MB curve, while the SIC-MF estimated curve is equally scattered but shifted upwards with respect to the SIC-MB curve (see enlarged views in the three insets in Fig.7c). In the first inset, it is clear that the SIC-MF curve could be shifted downwards simply by increasing the value of $\varepsilon$ in Eq.8.

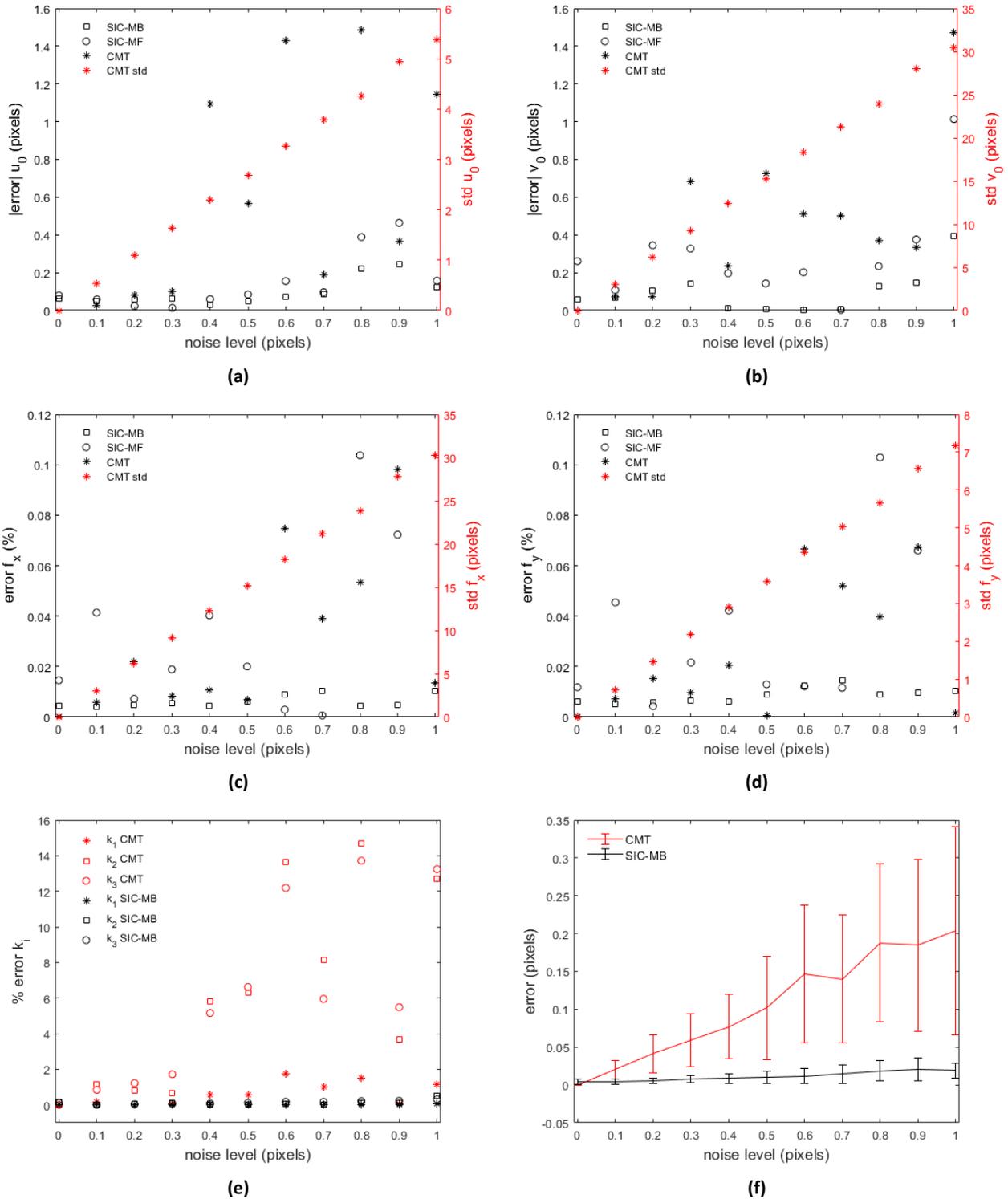

**Figure 8.** Comparison between the results of the CMT and SIC calibration for the principal point coordinates (a),(b), focal lengths (c),(d), and distortion parameters (e). Plot of the error calculated as the distance between the ground truth and the estimated undistorted points with the CMT and SIC-MB approaches (f).

*3.2 Real image data*

As reported in the previous section, in order to calibrate a camera from a single pose of the calibration target, it is essential to capture an image with a random distribution of grey levels covering the entire sensor area. For this purpose, a synthetic speckle image was generated by assigning a random value in the range [0, 255] to each pixel of a large image. The image was then displayed 1:1 on a Samsung SyncMaster 172B monitor ($1280 \times 1024\ pixels$ resolution, $0.264\ mm$ square pixel) placed at the camera working distance (~$500\ mm$, see pictures of the experimental setup in Fig.3 of the supplementary material, [26]). A consumer camera with a IMX179 Sony $3264 \times 2448\ pixels$ sensor and a $5 - 50\ mm$ varifocal lens was used for the experimental tests reported in this paper. First, the lens magnification was set so that the speckle image displayed on the monitor filled the entire FOV of the camera. Then, the speckle pattern was resized to obtain a captured image with an average speckle size of $3 \times 3\ pixels$. Finally, the original synthetic image was cropped to fit the monitor size and a Gaussian smoothing filter was applied. The position of any point in the 1:1 displayed speckle target is known with high accuracy by multiplying the pixel position in the synthetic image by the monitor pixel size, which is provided by the manufacturer with a resolution down to $10^{-3} mm$. An additional synthetic image with a grid of $30 \times 30\ pixels$ square markers with a $50\ pixels$ spacing was generated to serve as a traditional planar target (Fig.4, [26]). With the camera fixed to the optical table, the monitor was placed in ten different poses in the working volume and the two images of the speckle and of the dot patterns were sequentially displayed and captured by the camera. From the image of the entire dot grid, a subset of $13 \times 10$ of square markers was later selected to simulate a small planar target manually placed in multiple poses within the working volume as required by Zhang's method (Fig.9). A typical speckle image is shown in Figure 2d. The full set of 20 images (speckle and dot patterns) used in this work are provided as supplementary online data to allow the reproducibility of the results.

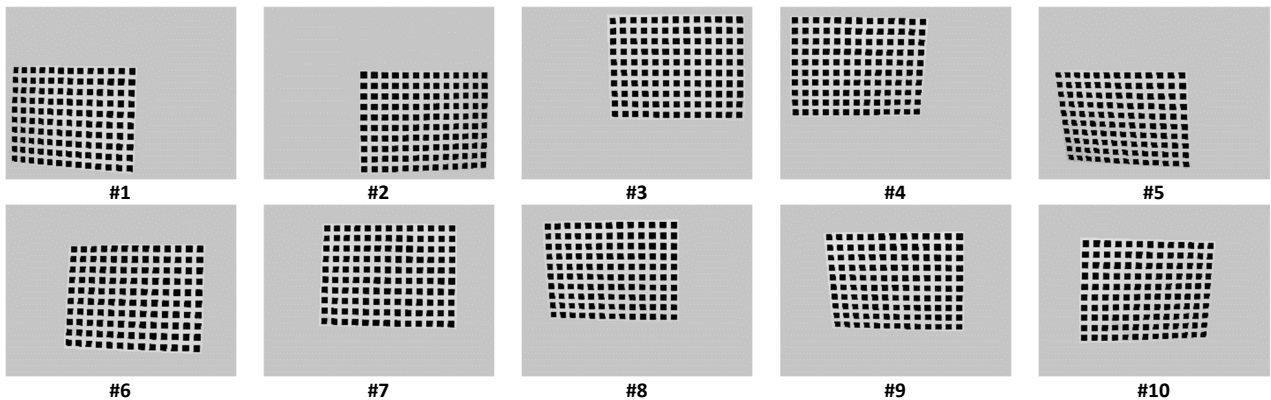

Figure 9. Images used for camera calibration using Zhang's method. In the same poses, the speckle pattern is displayed and each acquired image is processed separately with the SIC method (see typical speckle image in Fig.2d).

Each individual speckle image was processed using the SIC method. The captured image (Fig.2d) was used as the reference image for the DIC registration with the synthetic image (Fig.2c). A regular rectangular grid of points ($n_p = 117115$ points) was defined over the image, leaving an outer boundary equal in size to the DIC subset. The in-house developed DIC Matlab code described in [29] was used to perform the DIC correlation with a $21\ pixels$ subset, $8\ pixels$ spacing and a bicubic grey level interpolation. The DIC code features a SIFT [30] initialization and large deformation capabilities [29], both of which are required to accurately correlate images of different scale and large relative distortion (compare correlated grids in Fig.2c and Fig.2d). The resulting reference and current point grid coordinates represent the $\boldsymbol{p_d}$ and $\boldsymbol{p_w}$ points needed to start the SIC procedure (step #1, Fig.4). The points $\boldsymbol{p_d}$ are common to all acquired images, while a different set of target points $\boldsymbol{p_w}$ is obtained for each pose. Figure 10 shows the raw distribution of the total disparity obtained for the first image of the series at the beginning (panel a) and at the end (panel b) of the calibration process.

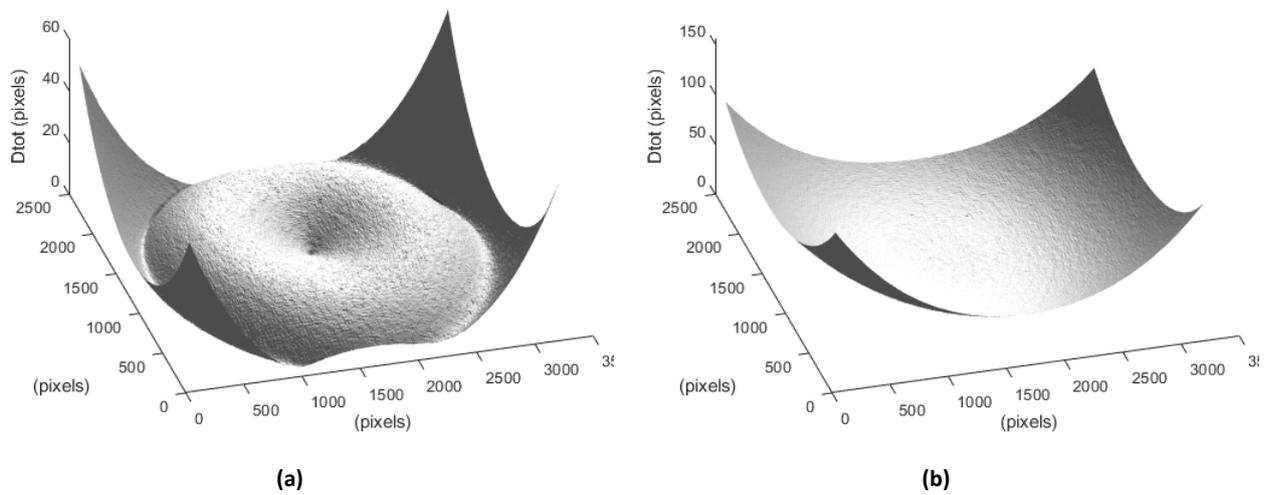

**Figure 10.** Total disparity map between distorted and reprojected points (raw data) obtained at the end of Step #1 for the real image in the pose #1 (a); total disparity map between distorted and undistorted points at the end of Step #3. Note the different scale for the z coordinate.

The raw radial distortion curves obtained with the SIC-MB and SIC-MF methods are shown in Fig.11 (where $r$ is the radial distance of the final undistorted points from the COD). For the sake of clarity, the curve obtained with CMT is not shown, as it overlaps the SCI-MB curve to a good approximation. The plots of the radial distortion curves for all the 10 images tested compared to the curve obtained with CMT are reported in Fig.5 of the supplementary material [26] for the two SIC approaches. The experimental data appear to be affected by a low level of noise (compare the curves in Fig.11 with those of the simulated data in Fig.7b). As for the simulated data, a shift between the curves of the two SIC approaches is observed due to the sensitivity to noise of the method used to calculate the scale factor $S$ (see enlarged views in the insets of Fig.11).

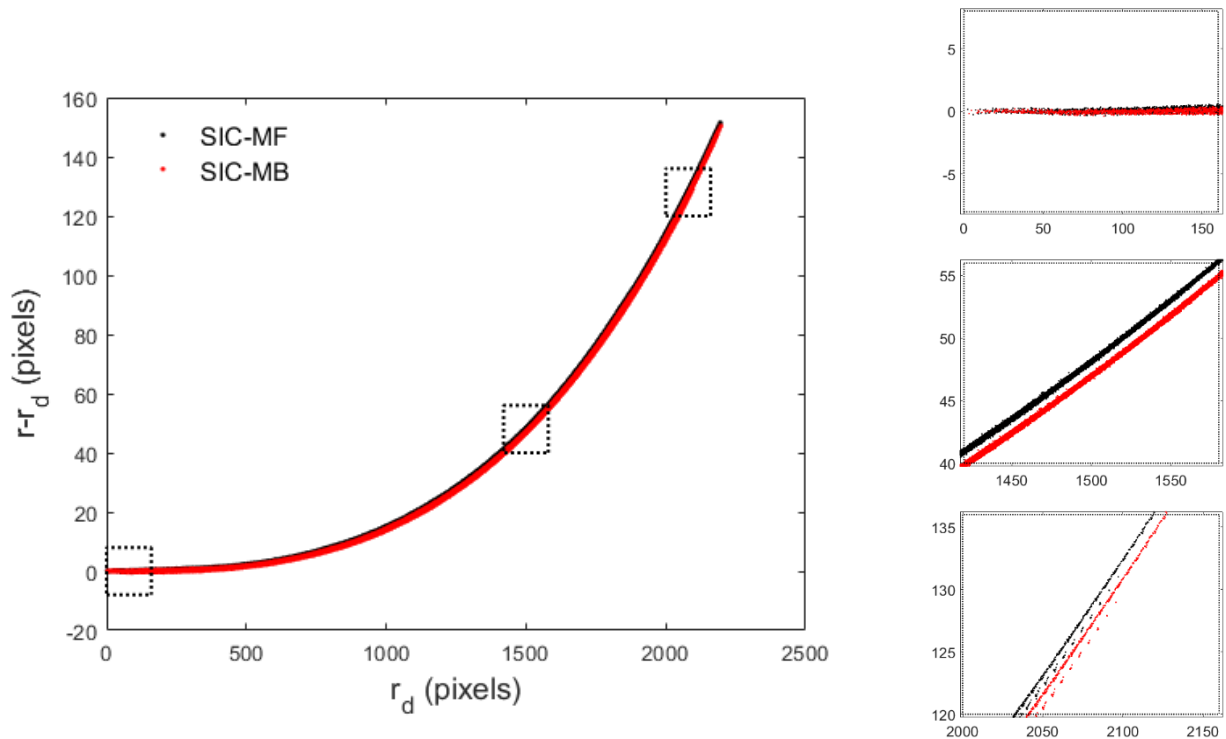

**Figure 11.** Comparison between the radial distortion curves obtained using the SIC-MF and SIC-MB approaches. The curve obtained with CMT is not shown as it overlaps with the SIC-MB curve with undetectable difference.

Table 2. Comparison between experimental results for the intrinsic parameters as obtained using CMT with 10 images of a dot grid pattern, and from each image of the speckle pattern processed separately using the proposed SIC method.

| | $CMT$ | $Im\ \#1$ | $Im\ \#2$ | $Im\ \#3$ | $Im\ \#4$ | $Im\ \#5$ | $Im\ \#6$ | $Im\ \#7$ | $Im\ \#8$ | $Im\ \#9$ | $Im\ \#10$ |
|---|---|---|---|---|---|---|---|---|---|---|---|
| $u_0$ (pixels) | 1707.37 | 1701.72 | 1713.80 | 1703.75 | 1712.15 | 1695.92 | 1704.58 | 1707.40 | 1704.19 | 1699.43 | 1716.99 |
| $v_0$ (pixels) | 1485.61 | 1496.64 | 1496.69 | 1492.44 | 1491.40 | 1485.95 | 1497.60 | 1497.61 | 1490.01 | 1487.03 | 1493.18 |
| $f_x$ (pixels) | 6618.29 | 6402.12 | 6928.39 | 6724.55 | 6632.96 | 6627.87 | 6513.80 | 6314.58 | 6648.16 | 6618.56 | 6628.84 |
| $f_y$ (pixels) | 6617.43 | 6410.37 | 6924.74 | 6718.45 | 6632.16 | 6620.14 | 6514.44 | 6314.24 | 6645.09 | 6611.16 | 6628.54 |
| $k_1$ | -0.623 | -0.576 | -0.672 | -0.632 | -0.609 | -0.612 | -0.596 | -0.560 | -0.613 | -0.607 | -0.609 |
| $k_2$ | 1.531 | 1.120 | 1.570 | 1.360 | 1.217 | 1.221 | 1.201 | 1.088 | 1.226 | 1.189 | 1.220 |
| $k_3$ | -5.475 | -3.011 | -5.394 | -4.194 | -3.674 | -3.454 | -3.375 | -2.998 | -3.549 | -3.338 | -3.695 |
| $RPE$ (pixels) | 0.15 ± 0.08 | 0.17 ± 0.10 | 0.17 ±0.12 | 0.17 ±0.11 | 0.21 ±0.17 | 0.21 ±0.13 | 0.18 ±0.11 | 0.16 ±0.10 | 0.19 ±0.13 | 0.21 ±0.13 | 0.21 ±0.17 |

Table 2 compares the intrinsic calibration results obtained from the ten images in Fig.9 using CMT with the results obtained from each individual speckle image in the same poses using the SIC-MB method. The results of the three approaches are plotted in Figure 12 as a function of the distance $t_z$ from $\Sigma_w$ to $\Sigma_c$ measured along optical axis of the lens (for brevity, *depth*). A clear trend can be seen in the plots. Depth-dependent distortion has long been studied and modeled in the literature [24][31][25][32]. However, the plots in Fig.12 are not sufficient to draw any conclusions about the possible relationship between distortion (and the associated focal length) and depth. In fact, in this study, measurement points in a generally oriented pose are not at the same $t_z$ from the camera frame (this also may be the cause of data dispersion). No clear relationship with depth was observed for the principal point coordinates (plots not shown).

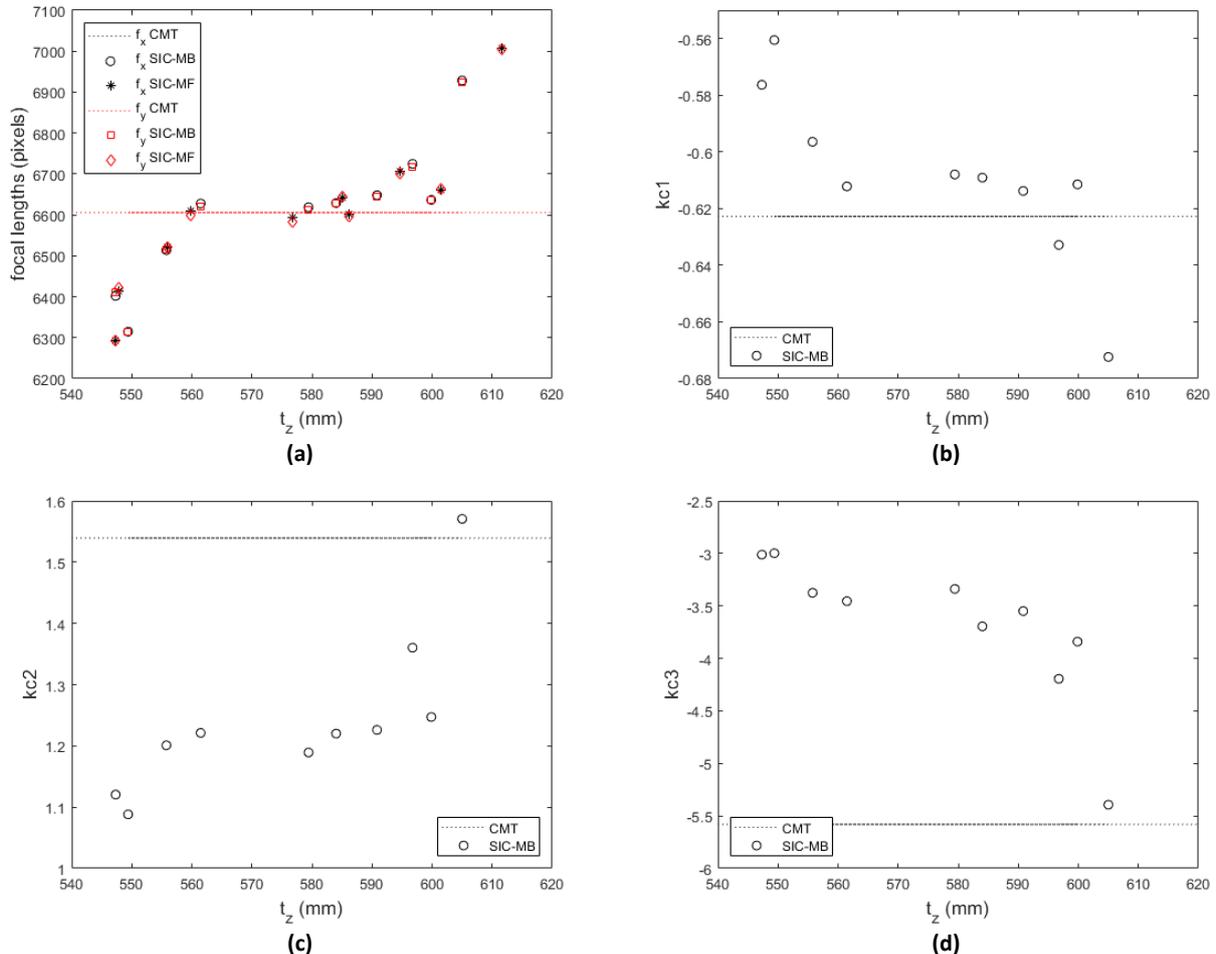

Figure 12. Plot of focal lengths vs depth as calculated by the three methods compared in this paper (a). Plots of the distortion coefficients vs depth as calculated by the SIC model-based approach and CMT (b) - (d).

The last row of Table 2 shows the value of the RPE for the CMT and for the SIC-MB methods. Note that the RPE for CMT refers to all points of all ten poses of the target as a result of the bundle adjustment optimization. Note, however, that although the target positions were distributed over the entire FOV, as required by Zhang's method, only a limited number of points are located in the outermost regions of the image (see the plot in Fig.6 in [26]). As a result, the distortion function is best fitted to the points in the most central region of the image, and the few peripheral points have only a modest effect on the total RPE. Interestingly, when the optimal homography evaluated with CMT for a given pose (e.g. pose #1 in Fig.9) is applied to all the markers of the original captured image (Fig.13a), it results in a higher value for the RPE ($0.47 \pm 0.34\ pixels$) and a non-radially symmetric distribution (Fig.13c). When tangential components are included in the distortion model, the RPE decreases to $0.33 \pm 0.27\ pixels$ with large values of RPE outside the target area (similar to Fig.13c). Conversely, with the SIC method, the measurement points are evenly distributed across the image up to the edges and a radial symmetric distribution of the RPE is obtained ($0.17 \pm 0.10\ pixels$, Fig.13d).

The plot in Figure 13d reveals an inhomogeneity of the RPE distribution, which is observed for all images tested in this work (plots not shown). This is most likely due to the flawed low quality lens/camera assembly used, and causes the radial distortion curve to split at large $r_d$ (see the last inset of Fig.11). This is a typical case where the use of a local model-free approach to distortion removal should be preferred to a global model-based approach.

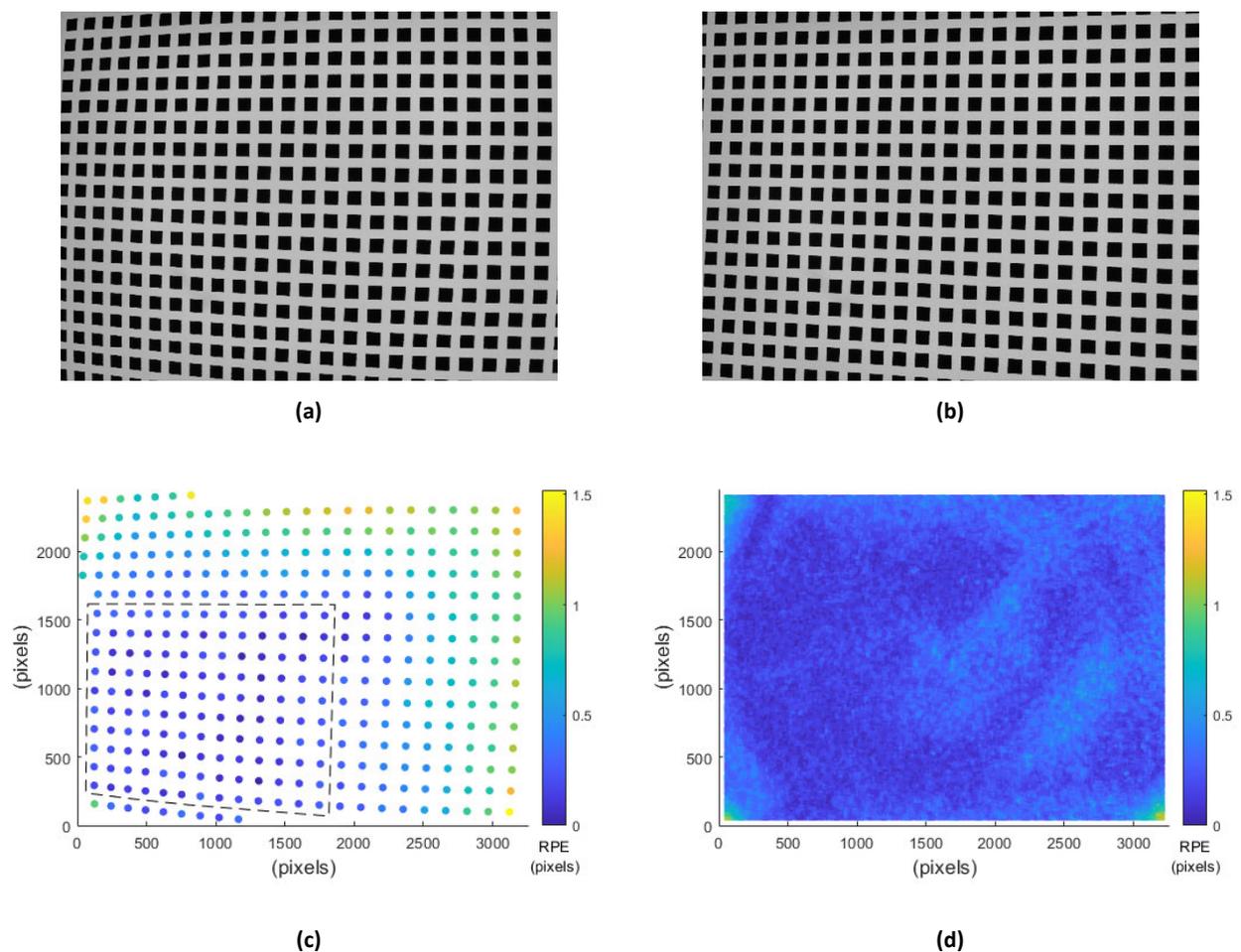

Figure 13. (a) Original and (b) undistorted image for the pose #1 computed using the SIC-MB method. (c) Plot of the reprojection error for all markers computed by the CMT method ($RPE = 0.47 \pm 0.34\ pixels$). The $13 \times 10$ dot grid used for CMT calibration is marked by a dashed line. (d) Plot of the reprojection error for the DIC grid as calculated by the SIC-MB method ($RPE = 0.17 \pm 0.10\ pixels$).

## 4. CONCLUDING REMARKS

This paper presents a method to accurately compute the complete set of camera parameters from a single image of a speckle pattern processed by DIC. The capabilities of the method are strictly related to the use of a dense and uniform grid of calibration points covering the entire sensor area. The method starts by evaluating the center of distortion. Although it is still quite common to assume that the COD coincides with the center of the sensor, the importance of determining the exact location of the COD for its effect on the other calibration parameters has long been recognized in the photogrammetry community [33][28]. To the best of the author's knowledge, the only metric calibration method that estimates the COD and the radial distortion curve from a single image has been proposed by Hartley and Kang [22]. However, the authors suggest using at least ten images of the target in different poses to obtain dense and robust results. Other COD estimation methods require special equipment and multiple images (e.g. [34][15]). Although the computation of the COD in this paper is iterative, the optimization problem is robust to noise and converges to the global optimum with an error < 1 pixel, regardless of the initial guess.

The proposed approach leads to an accurate (though slightly scaled) first estimate of the undistorted points by isolating a subset of distorted points centered in the evaluated COD, thus removing the effect that a rectangular sensor and a decentered COD have on the homography $H_d$. In fact, the reprojected points computed with the homography $H_d$ are *not* the ground truth, but an anisotropically scaled and translated copy of the ideal reprojected pattern. At the end of the Step #2 of the procedure, a very accurate estimate (error < 1%) of the full set of calibration parameters is obtained. In particular, intrinsic parameters are evaluated separately from the extrinsic parameters, thus avoiding possible coupling errors. The calibration can be further refined by optimization with either a model-based or a model-free distortion evaluation. Since the optimization starts with strict bounds from a configuration very close to the optimum, convergence to a local minimum is very unlikely.

Both simulated and real data demonstrate the accuracy and robustness of the proposed method. Obtained results disclose the limitations of (i) using the RPE as the sole indicator of calibration accuracy and (ii) assuming uniform calibration parameters across the working volume. The use of only one image of the calibration pattern may prove particularly useful in applications where the working volume is limited in size or physically constrained (e.g. universal testing machines, viewing through an inspection window, microscopes), as well as in the case of multi-camera systems, for which the calibration phase can be a long and tedious process.

A limitation of the proposed approach is the computational burden. In addition, the DIC algorithm requires large deformation analysis capabilities. Finally, the method is only applicable to close-range measurements as the FOV of the camera needs to be entirely filled by the monitor. The use of a phase speckle pattern may overcome the latter limitation and extend the use of SIC to large FOV measurements [35].